\documentclass[11pt,a4paper]{article}
\pdfoutput=1
\usepackage[hyperref]{acl2017}
\usepackage{times}
\usepackage{latexsym}

\usepackage{amsmath}
\usepackage{amsfonts}
\usepackage{amssymb}
\usepackage{booktabs}
\usepackage{multirow}
\usepackage{tabularx}
\usepackage{graphicx}
\usepackage{url}
\usepackage{algorithm}
\usepackage{algorithmicx,algpseudocode}
\usepackage{siunitx}
\usepackage{subcaption}
\aclfinalcopy 
\def\aclpaperid{598} 

\DeclareMathOperator*{\argmax}{arg\,max}

\newcommand{\E}{\mathbb{E}}
\newcommand{\Y}{\mathcal{Y}}

\title{Bandit Structured Prediction \\ for Neural Sequence-to-Sequence Learning}

\author{Julia Kreutzer$^{\ast}$ \and Artem Sokolov$^{\ast}$ \and Stefan Riezler$^{\dagger,\ast}$ \\
$^\ast$Computational Linguistics \& $^\dagger$IWR, Heidelberg University, Germany \\
{\tt \small\{kreutzer,sokolov,riezler\}@cl.uni-heidelberg.de} 
}

\date{}

\begin{document}

\maketitle

\begin{abstract}
Bandit structured prediction describes a stochastic optimization framework where learning is performed from partial feedback. This feedback is received in the form of a task loss evaluation to a predicted output structure, without having access to gold standard structures. 
We advance this framework by lifting linear bandit learning to neural sequence-to-sequence learning problems using attention-based recurrent neural networks.  Furthermore, we show how to incorporate control variates into our learning algorithms for variance reduction and improved generalization.  We present an evaluation on a neural machine translation task that shows improvements of up to 5.89 BLEU points for domain adaptation from simulated bandit feedback.   
\end{abstract}

\section{Introduction}

Many NLP tasks involve learning to predict a structured output such as a sequence, a tree or a graph. Sequence-to-sequence learning with neural networks  has recently become a popular approach that allows tackling structured prediction as a mapping problem between variable-length sequences, e.g., from foreign language sentences into target-language sentences \cite{SutskeverETAL:14}, or from natural language input sentences into linearized versions of syntactic \cite{VinyalsETAL:15} or semantic parses \cite{JiaLiang:16}.
A known bottleneck in structured prediction is the requirement of large amounts of gold-standard structures for supervised learning  of model parameters, especially for data-hungry neural network models. \citet{SokolovETAL:16,SokolovETALnips:16} presented a framework for stochastic structured prediction under bandit feedback that alleviates the need for labeled output structures in learning: Following an online learning protocol, on each iteration the learner receives an input, predicts an output structure, and receives partial feedback in form of a task loss evaluation of the predicted structure.\footnote{The name ``bandit feedback'' is inherited from the problem of maximizing the reward for a sequence of pulls of arms of so-called ``one-armed bandit'' slot machines.} They ``banditize'' several objective functions for linear structured predictions, and evaluate the resulting algorithms with simulated bandit feedback on various NLP tasks.

We show how to lift linear structured prediction under bandit feedback to non-linear models for sequence-to-sequence learning with attention-based recurrent neural networks \cite{BahdanauETAL:15}. Our framework is applicable to sequence-to-sequence learning from various types of weak feedback. For example, extracting learning signals from the execution of structured outputs against databases has been established in the communities of semantic parsing and grounded language learning since more than a decade \cite{ZettlemoyerCollins:05,ClarkeETAL:10,LiangETAL:11}.  Our work can build the basis for neural semantic parsing from weak feedback.

In this paper, we focus on the application of machine translation via neural sequence-to-sequence learning. The standard procedure of training neural machine translation (NMT) models is to compare their output to human-generated translations and to infer model updates from this comparison. However, the creation of reference translations or post-edits requires professional expertise of users. Our framework allows NMT models to learn from feedback that is weaker than human references or post-edits. One could imagine a scenario of personalized machine translation where translations have to be adapted to the user's specific purpose and domain. The feedback required by our methods can be provided by laymen users or can even be implicit, e.g., inferred from user interactions with the translated content on a web page.

Starting from the work of \citet{SokolovETAL:16,SokolovETALnips:16}, we lift their objectives to neural sequence-to-sequence learning. We evaluate the resulting algorithms on the task of French-to-English translation domain adaptation where a seed model trained on Europarl data is adapted to the NewsCommentary and the TED talks domain with simulated weak feedback. By learning from this feedback, we find 4.08 BLEU points improvements on NewsCommentary, and 5.89 BLEU points improvement on TED. 
Furthermore, we show how control variates can be integrated in our algorithms, yielding faster learning and improved generalization in our experiments.

\section{Related Work}

NMT models are most commonly trained under a word-level maximum likelihood objective. Even though this objective has successfully been applied to many sequence-to-sequence learning tasks, the resulting models suffer from exposure bias, since they learn to generate output words based on the history of given reference words, not on their own predictions. \citet{RanzatoETAL:16} apply techniques from reinforcement learning \cite{SuttonBarto:98,SuttonETAL:00} and imitation learning \cite{Schaal:99,RossETAL:11,DaumeETAL:09} to learn from feedback to the model's own predictions. Furthermore, they address the mismatch between word-level loss and sequence-level evaluation metric by using a mixture of the REINFORCE \cite{Williams:92} algorithm and the standard maximum likelihood training to directly optimize a sequence-level loss.
Similarly, \citet{ShenETAL:16} lift minimum risk training \cite{Och:03,SmithEisner:06,GimpelSmith:10,YuilleHe:12,HeDeng:12} from linear models for machine translation to NMT. These works are closely related to ours in that they use the technique of score function gradient estimators \cite{Fu:06,SchulmanETAL:15} for stochastic learning. However, the learning environment of \citet{ShenETAL:16} is different from ours in that they approximate the true gradient of the risk objective in a full information setting by sampling a subset of translations and computing the expectation over their rewards. In our bandit setting, feedback to only a single sample per sentence is available, making the learning problem much harder. The approach by \citet{RanzatoETAL:16} approximates the expectation with single samples, but still requires reference translations which are unavailable in the bandit setting.

To our knowledge, the only work on training NMT from weak feedback is the work by \citet{HeETAL:16}. They propose a dual-learning mechanism where two translation models are jointly trained on monolingual data. The feedback in this case is a reward signal from language models and a reconstruction error. This is attractive because the feedback can automatically be generated from monolingual data and does not require any human references. However, we are interested in using estimates of human feedback on translation quality to directly adapt the model to the users' needs.

Our approach follows most closely the work of \citet{SokolovETAL:16,SokolovETALnips:16}. They introduce bandit learning objectives for structured prediction and apply them to various NLP tasks, including machine translation with linear models. Their approach can be seen as an instantiation of reinforcement learning to one-state Markov decision processes under linear policy models. In this paper, we transfer their algorithms to non-linear sequence-to-sequence learning. \citet{SokolovETAL:16} showed applications of linear bandit learning to tasks such as multiclass-classification, OCR, and chunking, where learning can be done from scratch. We focus on lifting their linear machine translation experiments to the more complex NMT that requires a warm start for training. This is done by training a seed model on one domain and adapting it to a new domain based on bandit feedback only.
For this task we build on the work of \citet{FreitagETAL:16}, who investigate strategies to find the best of both worlds: models that adapt well to the new domain without deteriorating on the old domain. In contrast to previous approaches to domain adaptation for NMT, we do not require in-domain parallel data, but consult direct feedback to the translations generated for the new domain.

\section{Neural Machine Translation}

Neural models for machine translation are based on a sequence-to-sequence learning architecture consisting of an encoder and a decoder \cite{ChoETAL:14,SutskeverETAL:14,BahdanauETAL:15}. An encoder Recurrent Neural Network (RNN) reads in the source sentence and a decoder RNN generates the target sentence conditioned on the encoded source.

The input to the encoder is a sequence of vectors $\mathbf{x} = (x_1, \dots, x_{T_x})$ representing a sequence of source words of length $T_x$. In the approach of \citet{SutskeverETAL:14}, they are encoded into a single vector $c = q(\{ h_1, \dots, h_{T_x} \})$, where ${h_t = f(x_t, h_{t-1})}$ is the hidden state of the RNN at time $t$. Several choices are possible for the non-linear functions $f$ and $q$: Here we are using a Gated Recurrent Unit (GRU) \cite{ChungETAL:14} for $f$, and for $q$ an attention mechanism that defines the context vector as a weighted sum over encoder hidden states \cite{BahdanauETAL:15,LuongETAL:15}.

The decoder RNN predicts the next target word $y_t$ at time $t$ given the context vector $c$ and the previous target words $\mathbf{y}_{<t} = \{ y_1, \dots, y_{t-1} \}$ from a probability distribution over the target vocabulary $V$. This distribution is the result of a softmax transformation of the decoder outputs ${\mathbf{o} = \{o_1, \dots, o_{T_y}\}}$, such that
\begin{align*}
 p_{\theta} (y_t=w_i | \mathbf{y}_{<t}, c) = \frac{\exp(o_{w_i})}{\sum_{v=1}^{V}{\exp(o_{w_v})}}.
\end{align*}
The probability of a full sequence of outputs $\mathbf{y} = (y_1, \dots, y_{T_y})$ of length $T_y$ is defined as the product of the conditional word probabilities: 
\begin{align*}
p_{\theta}(\mathbf{y} | \mathbf{x}) = \prod^{T_y}_{t=1}{p_{\theta}(y_t | \mathbf{y}_{<t}, c)}.
\end{align*}
Since this encoder-decoder architecture is fully differentiable, it can be trained with gradient descent methods. Given a parallel training set of $S$ source sentences and their reference translations $D = \{(\mathbf{x}^{(s)}, \mathbf{y}^{(s)})\}^{S}_{s=1}$, we can define a word-level Maximum Likelihood Estimation (MLE) objective, which aims to find the parameters
\begin{align*}
  \hat{\mathbf{\theta}}^{\text{MLE}} =& \argmax_{\mathbf{\theta}}{ L^{\text{MLE}}(\mathbf{\theta})}
\end{align*}
of the following loss function:
\begin{align*}
L^{\text{MLE}}(\mathbf{\theta}) =& \sum^S_{s=1}{\log p_{\theta}(\mathbf{y}^{(s)} | \mathbf{x}^{(s)})}  \\
=&  \sum^S_{s=1}{\sum^{T_y}_{t=1}{\log p_{\theta}(y_t | \mathbf{x}^{(s)}, \mathbf{y}^{(s)}_{< t})}}.
\end{align*}
This loss function is non-convex for the case of neural networks. Clever initialization strategies, adaptive learning rates and momentum techniques are required to find good local maxima and to speed up convergence \cite{SutskeverETAL:13}. Another trick of the trade is to ensemble several models with different random initializations to improve over single models \cite{LuongETAL:15}.

At test time, we face a search problem to find the sequence of target words with the highest probability.
Beam search reduces the search error in comparison to greedy search, but also exponentially increases decoding time.

\begin{algorithm}[t]
  \label{alg:banditneural}
  \caption{Neural Bandit Structured Prediction}
  \begin{algorithmic}[1]
    \algnotext{EndFor}
    \Require Sequence of learning rates $\gamma_k$\label{alg:line:lrates:n}
    \Ensure Optimal parameters $\hat{\theta}$
    \State Initialize $\theta_0$ \label{alg:line:w0:n}
    \For{$k=0,\ldots,K$}
    \State Observe $\mathbf{x}_k$ \label{alg:line:input:n}
    \State Sample $\mathbf{\tilde y}_k \sim p_{\theta}(\mathbf{y}|\mathbf{x}_k)$ \label{alg:line:sample:n}
    \State Obtain feedback $\Delta(\mathbf{\tilde y}_k)$ \label{alg:line:feedback:n}
    \State $\theta_{k+1} = \theta_k - \gamma_k \; s_k$
    \EndFor
    \State Choose a solution $\hat{\theta}$ from the list $\{\theta_0, \ldots, \theta_K\}$
\end{algorithmic}
\end{algorithm}

\section{Neural Bandit Structured Prediction} 
\label{sec:neuralbandits}

Algorithm~1 is an adaptation of the Bandit Structured Prediction algorithm of \citet{SokolovETALnips:16} to neural models: For $K$ rounds, a model with parameters $\theta$ receives an input, samples an output structure, and receives user feedback. Based on this feedback, a stochastic gradient $s_k$ is computed and the model parameters are updated. As a post-optimization step, a solution $\hat{\theta}$ is selected from the iterates. This is done with online-to-batch conversion by choosing the model with optimal performance on held-out data. 

The core of the algorithm is the sampling: if the model distribution is very peaked, the model exploits, i.e., it presents the most probable outputs to the user. If the distribution is close to uniform, the model explores, i.e., it presents random outputs to the user. The balance between exploitation and exploration is crucial to the learning process: in the beginning the model is rather uninformed and needs to explore in order to find outputs with high reward, while in the end it ideally converges towards a peaked distribution that exactly fits the user's needs. Pre-training the model, i.e. setting $\theta_0$ wisely, ensures a reasonable exploitation-exploration trade-off. 

This online learning algorithm can be applied to any objective $L$ provided the stochastic gradients $s_k$ are unbiased estimators of the true gradient of the objective, i.e., we require $\nabla L = \E[s_k]$.
In the following, we will present objectives from \citet{SokolovETALnips:16} transferred to neural models, and explain how they can be enhanced by control variates.

\subsection{Expected Loss (EL) Minimization}

The first objective is defined as the expectation of a task loss $\Delta(\mathbf{\tilde y})$, e.g. $-\text{BLEU}(\mathbf{\tilde y})$, over all input and output structures:
\begin{align}
  \label{elobj}
  L^{\text{EL}}(\mathbf{\theta}) =& \E_{p(\mathbf{x})\,p_{\theta}({\mathbf{\tilde{y}}}|\mathbf{x})} \left[ \Delta(\mathbf{\tilde y}) \right].
\end{align}
In the case of full-information learning where reference outputs are available, we could evaluate all possible outputs against the reference to obtain an exact estimation of the loss function. However, this is not feasible in our setting since we only receive partial feedback for a single output structure per input. Instead, we use stochastic approximation to optimize this loss. The stochastic gradient for this objective is computed as follows:
\begin{align}
  \label{sg-elobj}
s^{\text{EL}}_k =&  \Delta(\mathbf{\tilde{y}}) \frac{\partial \log p_{\theta}(\mathbf{\tilde{y}}|\mathbf{x}_k)}{\partial \mathbf{\theta}}. 
\end{align}
Objective \eqref{elobj} is known from minimum risk training \cite{Och:03} and has been lifted to NMT by \citet{ShenETAL:16} -- but not for learning from weak feedback. Equation \eqref{sg-elobj} is an instance of the score function gradient estimator \cite{Fu:06} where
\begin{align}
  \label{scorefunction}
  \nabla \log p_{\theta}(\mathbf{\tilde{y}}|\mathbf{x}_k)
  \end{align} denotes the score function. We give an algorithm to sample structures from an encoder-decoder model in Algorithm~2. It corresponds to the algorithm presented by \citet{ShenETAL:16} with the difference that it samples single structures, does not assume a reference structure, and additionally returns the sample probabilities. A similar objective has also been used in the REINFORCE algorithm \cite{Williams:92} which has been adapted to NMT by \citet{RanzatoETAL:16}. 

\subsection{Pairwise Preference Ranking (PR) }
\label{sec:pw}

The previous objective requires numerical feedback as an estimate of translation quality. Alternatively, we can learn from pairwise preference judgments that are formalized in preference ranking objectives.
Let ${\mathcal{P}(\mathbf{x}) = \{ \left< \mathbf{y}_i,\mathbf{y}_j \right> | \mathbf{y}_i, \mathbf{y}_j \in \Y(\mathbf{x})\}}$ denote the set of output pairs for an input $\mathbf{x}$, and let $\Delta(\left< \mathbf{y}_i,\mathbf{y}_j \right>): \mathcal{P}(\mathbf{x}) \rightarrow [0,1]$ denote a task loss function that specifies a dispreference of $y_i$ over $y_j$. In our experimental simulations we use two types of pairwise feedback. Firstly, continuous pairwise feedback\footnote{Note that our definition of continuous feedback is slightly different from the one proposed in \citet{SokolovETALnips:16} where updates are only made for misrankings.} is computed as
\begin{align*}
  \Delta( \left< \mathbf{y}_i,\mathbf{y}_j \right> ) =  \Delta(\mathbf{y}_j) - \Delta(\mathbf{y}_i),
\end{align*}
and secondly, binary feedback is computed as
\begin{align*}
  \Delta( \left< \mathbf{y}_i,\mathbf{y}_j \right> ) =
\begin{cases}
  1  \quad \text{if } \Delta(\mathbf{y}_j) > \Delta(\mathbf{y}_i),\\
  0 \quad \text{otherwise.}
\end{cases}
\end{align*}
Analogously to the sequence-level sampling for linear models \cite{SokolovETALnips:16}, we define the following probabilities for word-level sampling:
\begin{align*}
  p_{\theta}^+(\tilde y_t = w_i |\mathbf{x}, \mathbf{\hat y}_{< t}) &= \frac{\exp(o_{w_i})}{\sum_{v=1}^{V}{\exp(o_{w_v})}},\\
  p_{\theta}^-(\tilde y_t = w_j |\mathbf{x}, \mathbf{\hat y}_{< t}) &= \frac{\exp(-o_{w_j})}{\sum_{v=1}^{V}{\exp(-o_{w_v})}}.
\end{align*}
The effect of the negation within the softmax is that the two distributions $p_{\theta}^+$ and $p_{\theta}^-$ rank the next candidate target words $\tilde{y}_t$ (given the same history, here the greedy output $\mathbf{\hat y}_{< t}$) in opposite order. Globally normalized models as in the linear case, or LSTM-CRFs \cite{HuangETAL:15} for the non-linear case would allow sampling full structures such that the ranking over full structures is reversed. But in the case of locally normalized RNNs we retrieve only locally reversed-rank samples.
Since we want the model to learn to rank $\mathbf{\tilde y}_i$ over $\mathbf{\tilde y}_j$, we would have to sample $\mathbf{\tilde y}_i$ word-by-word from $p^+_{\theta}$ and $\mathbf{\tilde y}_j$ from $p^-_{\theta}$. However, sampling all words of $\mathbf{\tilde y}_j$ from $p_{\theta}^-$ leads to translations that are neither fluent nor source-related, so we propose to randomly choose one position of $\mathbf{\tilde y}_j$ where the next word is sampled from $p^-_{\theta}$ and sample the remaining words from $p^+_{\theta}$. We found that this method produces suitable negative samples, which are only slightly perturbed and still relatively fluent and source-related. A detailed algorithm is given in Algorithm~3. 

In the same manner as for linear models, we define the probability of a pair of sequences as
\begin{align*}
p_{\theta}(\left<\mathbf{\tilde y}_i, \mathbf{\tilde y}_j\right>|\mathbf{x}) &= p^+_{\theta}( \mathbf{\tilde y}_i|\mathbf{x}) \times p^-_{\theta}(\mathbf{\tilde y}_j|\mathbf{x}).
\end{align*}
Note that with the word-based sampling scheme described above, the sequence $\mathbf{\tilde y}_j$ also includes words sampled from $p^+_{\theta}$.\\
The pairwise preference ranking objective expresses an expectation over losses over these pairs:
\begin{align}
  \label{pwel}
  L^{\text{PR}}(\mathbf{\theta}) =& \E_{p(\mathbf{x})\,p_{\theta}(\left<\mathbf{\tilde y}_i, \mathbf{\tilde y}_j\right>|\mathbf{x})} \left[ \Delta(\left<\mathbf{\tilde y}_i, \mathbf{\tilde y}_j\right>) \right].
\end{align}
The stochastic gradient for this objective is
\begin{align}
  \label{sg-pwel}
s_k^{\text{PR}} =& \Delta(\left<\mathbf{\tilde y}_i, \mathbf{\tilde y}_j\right>) \\ \notag
& \times \left(\frac{\partial \log p^+_{\theta}(\mathbf{\tilde y}_i|\mathbf{x}_k)}{\partial \mathbf{\theta}}  +\frac{\partial \log p^-_{\theta}(\mathbf{\tilde y}_j|\mathbf{x}_k)}{\partial \mathbf{\theta}} \right). \\ \notag
\end{align}
This training procedure resembles well-known approaches for noise contrastive estimation \cite{GutmannETAL:10} with negative sampling that are commonly used for neural language modeling \cite{CollobertETAL:11,MnihTeh:12,MikolovETAL:13}. In these approaches, negative samples are drawn from a non-parametric noise distribution, whereas we draw them from the perturbed model distribution.

\begin{algorithm}[t]
	\label{alg:sampling}
		\caption{Sampling Structures}
		\begin{algorithmic}[1]
			\Require{Model $\mathbf{\theta}$, target sequence length limit $T_y$}
			\Ensure{Sequence of words ${\mathbf{w}=(w_1, \dots, w_T{_y})}$ and log-probability $p$}
			 \State{$w_0 = \text{START}$, $p_0 = 0$}
			 \State{$\mathbf{w} = (w_0)$}
			 \For{$t \leftarrow 1 \dots T_y$} 
			 		\State{$w_t \sim p_{\theta}(w | \mathbf{x}, \mathbf{w}_{<t}) $}
			 			\State{$p_t = p_{t-1} + \log p_{\theta}(w | \mathbf{x}, \mathbf{w}_{<t})$}
			 			\State{$\mathbf{w} = (w_1, \dots, w_{t-1}, w_t)$} 
			 \EndFor
		\State{Return $\textbf{w}$ and $p_T$}
		\end{algorithmic}
	\end{algorithm}
		
\begin{algorithm}[t]
	\label{alg:negsampling}
		\caption{Sampling Pairs of Structures}
		\begin{algorithmic}[1]
			\Require{Model $\mathbf{\theta}$, target sequence length limit $T_y$}
			\Ensure{Pair of sequences $\left<\mathbf{w}, \mathbf{w}'\right>$} and their log-probability $p$
			 \State{$p_0 = 0$}
			 \State{$\mathbf{w}, \mathbf{w'}, \mathbf{\hat{w}} = (\text{START})$}
			 \State{$i \sim \mathcal{U}(1, T)$}
			 \For{$t \leftarrow 1 \dots T_y$} 
			 		\State{$\hat{w}_t = \arg \max_{w \in V} p^+_{\theta}(w | \mathbf{x}, \mathbf{\hat{w}}_{<t}) $} 
			 		\State{$w_t \sim p^+_{\theta}(w | \mathbf{x}, \mathbf{\hat{w}}_{<t}) $} 
			 		\State{$p_t = p_{t-1} + \log p^+_{\theta}(w_t | \mathbf{x}, \mathbf{\hat{w}}_{<t})$ }
			 		\If{$i=t$} 
			 			\State{$w'_t \sim p^-_{\theta}(w | \mathbf{x}, \mathbf{\hat{w}}_{<t}) $} 
			 			\State{$p_t = p_t + \log p^-_{\theta}(w'_t | \mathbf{x}, \mathbf{\hat{w}}_{<t})$ }
			 		\Else
			 			\State{$w'_t \sim p^+_{\theta}(w | \mathbf{x}, \mathbf{\hat{w}}_{<t}) $}
			 			\State{$p_t = p_t + \log p^+_{\theta}(w'_t | \mathbf{x}, \mathbf{\hat{w}}_{<t})$ }
			 		\EndIf
			 		
			 		\State{$\mathbf{w} = (w_1, \dots, w_{t-1}, w_t)$} 
			 		\State{$\mathbf{w'} = (w'_1, \dots, w'_{t-1}, w'_t)$} 
			 		\State{$\mathbf{\hat{w}} = (\hat{w}_1, \dots, \hat{w}_{t-1}, \hat{w}_t)$} 
			 \EndFor
		\State{Return $\left<\mathbf{w}, \mathbf{w}'\right>$ and $p_T$}
		\end{algorithmic}
	\end{algorithm}

\subsection{Control Variates}

The stochastic gradients defined in equations \eqref{sg-elobj} and \eqref{sg-pwel} can be used in stochastic gradient descent optimization \cite{BottouETAL:16} where the full gradient is approximated using a minibatch or a single example in each update. The stochastic choice, in our case on inputs and outputs, introduces noise that leads to slower convergence and degrades performance. In the following, we explain how antithetic and additive control variate techniques from Monte Carlo simulation \cite{Ross:13} can be used to remedy these problems.

The idea of additive control variates is to augment a random variable $X$ whose expectation is sought, by another random variable $Y$ to which $X$ is highly correlated. $Y$ is then called the control variate. Let $\bar{Y}$ furthermore denote its expectation. Then the following quantity $X-\hat{c}\,Y + \hat{c}\,\bar{Y}$ is an unbiased estimator of $\E[X]$. In our case, the random variable of interest is the noisy gradient $X = s_k$ from Equation \eqref{sg-elobj}. The variance reduction effect of control variates can be seen by computing the variance of this quantity:
\begin{align}
\label{eq:var-controlvariate} 
\text{Var}(X-\hat{c}\,Y) = & \; \text{Var}(X) + \hat{c}^2\text{Var}(Y) \\ \notag
& - 2\hat{c}\,\text{Cov}(X,Y).
\end{align}
Choosing a control variate such that $\text{Cov}(X,Y)$ is positive and high enough, the variance of the gradient estimate will be reduced. 

An example is the average reward baseline known from reinforcement learning \cite{Williams:92}, yielding 
\begin{align}
\label{eq:baseline-controlvariate} 
Y_k = \nabla \log p_{\theta}(\mathbf{\tilde{y}}|\mathbf{x}_k) \, \frac{1}{k}\sum_{j=1}^k \Delta(\mathbf{\tilde y}_j).
\end{align}
The optimal scalar $\hat{c}$ can be derived easily by taking the derivative of \eqref{eq:var-controlvariate}, leading to $\hat{c} = \frac{\text{Cov}(X,Y)}{\text{Var}(X)}.$
This technique has been applied to using the score function (Equation \eqref{scorefunction}) as control variate in \citet{RanganathETAL:14}, yielding the following control variate:
\begin{align}
\label{eq:scorefunction-controlvariate} 
Y{_k} = \nabla \log p_{\theta}(\mathbf{\tilde{y}}|\mathbf{x}_k).
\end{align}
Note that for both types of control variates, \eqref{eq:baseline-controlvariate} and \eqref{eq:scorefunction-controlvariate}, the expectation $\bar{Y}$ is zero, simplifying the implementation. However, the optimal scalar $\hat{c}$ has to be estimated for every entry of the gradient separately for the score function control variate. We will explore both types of control variates for the stochastic gradient \eqref{sg-elobj} in our experiments.

A further effect of control variates is to reduce the magnitude of the gradient, the more so the more the stochastic gradient and the control variate covary. For $L$-Lipschitz continuous functions, a reduced gradient norm directly leads to a bound on $L$ which appears in the algorithmic stability bounds of \citet{HardtETAL:16}. This effect of improved generalization by control variates is empirically validated in our experiments.

A similar variance reduction effect can be obtained by antithetic control variates. Here $\E[X]$ is approximated by the estimator $\frac{X_1 + X_2}{2}$ whose variance is
\begin{align}
\label{eq:var-antithetic} 
\text{Var}\left(\frac{X_1 + X_2}{2}\right) & = \frac{1}{4} \big(\text{Var}(X_1)  \\ \notag
&+ \text{Var}(X_2) + 2\text{Cov}(X_1,X_2) \big). \notag
\end{align}
Choosing the variates $X_1$ and $X_2$ such that $\text{Cov}(X_1,X_2)$ is negative will reduce the variance of the gradient estimate. Under certain assumptions, the stochastic gradient \eqref{sg-pwel} of the pairwise preference objective can be interpreted as an antithetic estimator of the score function \eqref{scorefunction}. The antithetic variates in this case would be
\begin{align}
\label{eq:pairwise-antithetic} 
X{_1} & = \nabla \log p^+_{\theta}(\mathbf{\tilde y}_i|\mathbf{x}_k), \\ \notag
X{_2} & = \nabla \log p^-_{\theta}(\mathbf{\tilde y}_j|\mathbf{x}_k), \notag
\end{align}
where an antithetic dependence of $X_2$ on $X_1$ can be achieved by construction of $p^+_{\theta}$ and $p^-_{\theta}$ (see \citet{Capriotti:08} which is loosely related to our approach). Similar to control variates, antithetic variates have the effect of shrinking the gradient norm, the more so the more the variates are antithetically correlated, leading to possible improvements in algorithmic stability \cite{HardtETAL:16}.

\section{Experiments}
In the following, we present an experimental evaluation of the learning objectives presented above on machine translation domain adaptation. We compare how the presented neural bandit learning objectives perform in comparison to linear models, then discuss the handling of unknown words and eventually investigate the impact of techniques for variance reduction.

\subsection{Setup}

\paragraph{Data.} We perform domain adaptation from Europarl (EP) to News Commentary (NC) and TED talks (TED) for translations from French to English.
Table \ref{tab:data} provides details about the datasets. For data pre-processing we follow the procedure of \citet{SokolovETAL:16,SokolovETALnips:16} using \texttt{cdec} tools for filtering, lowercasing and tokenization.
The challenge for the bandit learner is to adapt from the EP domain to NC or TED with weak feedback only. 

\begin{table}[t]
\begin{center}
\resizebox{0.9\columnwidth}{!}{
\begin{tabular}{ll|lll}
\toprule
\bf Domain &\bf Version &\bf Train &\bf Valid. &\bf Test\\
\midrule

Europarl & v.5 & 1.6M & 2k & 2k\\
News Commentary & WMT07 & 40k & 1k & 2k\\ 
TED & TED2013 & 153k & 2k & 2k\\
\bottomrule
\end{tabular}
}
\end{center}
\caption{Number of parallel sentences for training, validation and test sets for French-to-English domain adaptation.}
\label{tab:data}
\end{table}

\paragraph{NMT Models.}
We choose a standard encoder-decoder architecture with single-layer GRU RNNs with 800 hidden units, a word embedding size of 300 and $\tanh$ activations. The encoder consists of a bidirectional RNN, where the hidden states of backward and forward RNN are concatenated. The decoder uses the attention mechanism proposed by \citet{BahdanauETAL:15}.\footnote{We do not use beam search nor ensembling, although we are aware that higher performance is almost guaranteed with these techniques. Our goal is to show relative differences between different models, so a simple setup is sufficient for the purpose of our experiments.} 
Source and target vocabularies contain the 30k most frequent words of the respective parts of the training corpus. We limit the maximum sentence length to 50. Dropout \cite{srivastava2014dropout} with a probability of 0.5 is applied to the network in several places: on the embedded inputs, before the output layer, and on the initial state of the decoder RNN.
The gradient is clipped when its norms exceeds 1.0 to prevent exploding gradients and stabilize learning \cite{pascanu2013difficulty}. All models are implemented and trained with the sequence learning framework \texttt{Neural Monkey} \cite{libovicky2016cuni, bojar2016ufal}. \footnote{The \texttt{Neural Monkey} fork \url{https://github.com/juliakreutzer/bandit-neuralmonkey} contains bandit learning objectives and the configuration files for our experiments.} They are trained with a minibatch size of 20, fitting onto single 8GB GPU machines. The training dataset is shuffled before each epoch.

\paragraph{Baselines.}
The out-of-domain baseline is trained on the EP training set with standard MLE. For both NC and TED domains, we train two full-information in-domain baselines: The first in-domain baseline is trained on the relatively small in-domain training data. The second in-domain baseline starts from the out-of-domain model and 
is further trained on the in-domain data.
All baselines are trained with MLE and Adam \cite{kingma2014adam} ($\alpha = \num{1e-4}$, $\beta_1 = 0.9$, $\beta_2 = 0.999$) until their performance stops increasing on respective held-out validation sets. 
The gap between the performance of the out-of-domain model and the in-domain models defines the range of possible improvements for bandit learning. All models are evaluated with Neural Monkey's \texttt{mteval}. For statistical significance tests we used Approximate Randomization testing \cite{Noreen:89}. 

\paragraph{Bandit Learning.}
Bandit learning starts with the parameters of the out-of-domain baseline. The bandit models are expected to improve over the out-of-domain baseline by receiving feedback from the new domain, but at most to reach the in-domain baseline since the feedback is weak. 
The models are trained with Adam on in-domain data for at most 20 epochs.
Adam's step-size parameter $\alpha$  was tuned on the validation set and was found to perform best when set to \num{1e-5} for non-pairwise, \num{1e-6} for pairwise objectives on NC, \num{1e-7} for pairwise objectives on TED.
The best model parameters, selected with early stopping on the in-domain validation set, are evaluated on the held-out in-domain test set. 
In the spirit of \citet{FreitagETAL:16} they are additionally evaluated on the out-of-domain test set to investigate how much knowledge of the old domain the models lose while adapting to the new domain.
Bandit learning experiments are repeated two times, with different random seeds, and mean BLEU scores with standard deviation are reported. 

\paragraph{Feedback Simulation.}
Weak feedback is simulated from the target side of the parallel corpus, but references are never revealed to the learner.
\citet{SokolovETAL:16,SokolovETALnips:16} used a smoothed version of per-sentence BLEU for simulating the weak feedback for generated translations from the comparison with reference translations. Here, we use gGLEU instead, which \citet{wu2016google} recently introduced for learning from sentence-level reward signals correlating well with corpus BLEU. This metric is closely related to BLEU, but does not have a brevity penalty and considers the recall of matching $n$-grams. It is defined as the minimum of recall and precision over the total $n$-grams up to a certain $n$. 
Hence, for our experiments $\Delta(\mathbf{\tilde{y}}) = -\text{gGLEU}(\mathbf{\tilde{y}}, \mathbf{y})$, where $\mathbf{\tilde{y}}$ is a sample translation and $\mathbf{y}$ is the reference translation.

\paragraph{Unknown words.}
One drawback of NMT models is their limitation to a fixed source- and target vocabulary. In a domain adaptation setting, this limitation has a critical impact to the translation quality. The larger the distance between old and new domain, the more words in the new domain are unknown to the models trained on the old domain (represented with a special UNK token). We consider two strategies for this problem for our experiments: 
\begin{enumerate}
\item UNK-Replace: \citet{jean-EtAl:2015:WMT} and \citet{luong-EtAl:2015:ACL-IJCNLP} replace generated UNK tokens with aligned source words or their lexical translations in a post-processing step. \citet{FreitagETAL:16} and \citet{hashimoto-eriguchi-tsuruoka:2016:WAT2016} demonstrated that this technique is beneficial for NMT domain adaptation.
\item BPE: \citet{sennrich-haddow-birch:2016:P16-12} introduce byte pair encoding (BPE) for word segmentation to build translation models on sub-word units. Rare words are decomposed into subword units, while the most frequent words remain single vocabulary items.
\end{enumerate} 
For UNK-Replace we use \texttt{fast\_align} to generate lexical translations on the EP training data. When an UNK token is generated, we look up the attention weights and find the source token that receives most attention in this step. If possible, we replace the UNK token by its lexical translation. If it is not included in the lexical translations, it is replaced by the source token. The main benefit of this technique is that it deals well with unknown named entities that are just passed through from source to target. However, since it is a non-differentiable post-processing step, the NMT model cannot directly be trained for this behavior.
Therefore we also train sub-word level NMT with BPE. We apply 29,800 merge operations to obtain a vocabulary of 29,908 sub-words. The procedure for training these models is exactly the same as for the word-based models. The advantage of this method is that the model is in principle able to generate any word composing it from sub-word units. However, training sequences become longer and candidate translations are sampled on a sub-word level, which introduces the risk of sampling nonsense words. 

\paragraph{Control variates.} We implement the average baseline control variate as defined in Equation~7, which results in keeping an running average over previous losses. Intuitively, absolute gGLEU feedback is turned into relative feedback that reflects the current state of the model. The sign of the update is switched when the gGLEU for the current sample is worse than the average gGLEU, so the model makes a step away from it, while in the case of absolute feedback it would still make a small step towards it. In addition, we implement the score function control variate with a running estimate ${\hat{c}_k = \frac{1}{k} \sum_{j=1}^{k}{\frac{\text{Cov}(s_j,\nabla \log p_{\theta}(\mathbf{\tilde{y}}_j|\mathbf{x}_j))}{\text{Var}(s_j)}}}$.

\begin{table}[t]
\begin{center}
\resizebox{1\columnwidth}{!}{
\begin{tabular}{l|ll|l|l|l}
\toprule
\textbf{Algorithm}  & \textbf{Train data} & \textbf{Iter.} & \textbf{EP} & \textbf{NC} & \textbf{TED} \\
\midrule
MLE &  EP & 12.3M & 31.44 & 26.98 & 23.48\\
MLE-UNK &   &         & 31.82 & 28.00 & 24.59\\
MLE-BPE &   &     12.0M  & 31.81 & 27.20 & 24.35\\
\bottomrule
\end{tabular}
}
\end{center}
\caption{Out-of-domain NMT baseline results (BLEU) on in- and out-of-domain test sets trained only on EP data.}
\label{tab:nmtbaselines:out}
\end{table}

\begin{table}[t]
\begin{center}
\resizebox{1\columnwidth}{!}{
\begin{tabular}{l|ll|l|l}
\toprule
\textbf{Algorithm}  & \textbf{Train data} & \textbf{Iter.} & \textbf{EP} & \textbf{NC}  \\
\midrule
MLE  & NC & 978k  & 13.67 & 22.32 \\ 
MLE-UNK & &           & 13.83  &  22.56\\ 
MLE-BPE & &       1.0M    & 14.09 & 23.01\\
\midrule
MLE &  EP$\rightarrow$NC & 160k & 26.66 & 31.91  \\ 
MLE-UNK & & & 27.19 & 33.19  \\
MLE-BPE & & 160k & 27.14 & 33.31  \\
\toprule
\textbf{Algorithm}  & \textbf{Train data} & \textbf{Iter.} & \textbf{EP} & \textbf{TED} \\
\midrule
MLE  & TED & 2.2M  & 14.16 &  32.71\\ 
MLE-UNK & &           & 15.15 & 33.16\\ 
MLE-BPE & &  3.0M     & 14.18 & 32.81\\ 
\midrule
MLE &  EP$\rightarrow$TED & 460k & 23.88 &  33.65\\ 
MLE-UNK &   & & 24.64 &  35.57 \\
MLE-BPE &   & 2.2M & 23.39  & 36.23 \\
\bottomrule
\end{tabular}
}
\end{center}
\caption{In-domain NMT baselines results (BLEU) on in- and out-of-domain test sets. 
The EP$\rightarrow$NC is first trained on EP, then fine-tuned on NC. The EP$\rightarrow$TED is first trained on EP, then fine-tuned on TED.}
\label{tab:nmtbaselines:in}
\end{table}

\begin{table*}[t]
\begin{center}

\resizebox{\textwidth}{!}{%

\begin{subtable}{0.66\textwidth}
\flushleft
\begin{tabular}[t]{l|l|l||l|l}
\toprule
\textbf{Algorithm}   & \textbf{Iter.} & \textbf{EP} & \textbf{NC} & \textbf{Diff.}\\
\toprule  
EL & 317k & 30.36$_{\pm 0.20}$& 29.34$_{\pm 0.29}$ & 2.36\\ 
EL-UNK* & 317k &  30.73$_{\pm 0.20}$& 30.33$_{\pm 0.42}$ & 2.33 \\ 
EL-UNK**  & 349k &  30.67$_{\pm 0.04}$ &  30.45$_{\pm 0.27}$ & 2.45\\ 
EL-BPE  & 543k & 30.09$_{\pm 0.31}$  & 30.09$_{\pm 0.01}$ & 2.89\\ 
\midrule
PR-UNK** (bin)   & 22k & 30.76$_{\pm 0.03}$ & 29.40$_{\pm 0.02}$ & 1.40\\ 
PR-BPE (bin) & 14k & 31.02$_{\pm 0.09}$ & 28.92$_{\pm 0.03}$ & 1.72\\
PR-UNK** (cont) & 12k & 30.81$_{\pm 0.02}$ & 29.43$_{\pm 0.02}$ & 1.43 \\ 
PR-BPE (cont) & 8k  & 30.91$_{\pm 0.01}$ & 28.99$_{\pm 0.00}$ & 1.79\\ 
\midrule
SF-EL-UNK**  & 713k & 29.97$_{\pm 0.09}$ & 30.61$_{\pm 0.05}$ & 2.61\\
SF-EL-BPE &  375k & 30.46$_{\pm 0.10}$  & 30.20$_{\pm 0.11}$ & 3.00\\ 
\midrule
BL-EL-UNK** & 531k & 30.19$_{\pm 0.37}$ & 31.47$_{\pm 0.09}$ & 3.47\\ 
BL-EL-BPE  & 755k & 29.88$_{\pm 0.07}$ & 31.28$_{\pm 0.24}$ & \textbf{4.08} \\ 
\bottomrule
\end{tabular}
\subcaption{Domain adaptation from EP to NC.}
\end{subtable}

\begin{subtable}{0.66\textwidth}
\flushright
\begin{tabular}[t]{l|l|l||l|l}
\toprule
\textbf{Algorithm}  & \textbf{Iter.} & \textbf{EP} & \textbf{TED} & \textbf{Diff.}\\
\toprule 
EL   &  976k & 29.34$_{\pm 0.42}$ & 27.66$_{\pm 0.03}$ & 4.18\\ 
EL-UNK*   & 976k &  29.68$_{\pm 0.29}$ & 29.44$_{\pm 0.06}$ & 4.85\\ 
EL-UNK**  & 1.1M & 29.62$_{\pm 0.15}$ & 29.77$_{\pm 0.15}$ & 5.18\\ 
EL-BPE  & 831k &  30.03$_{\pm 0.43}$ & 28.54$_{\pm 0.04}$ & 4.18\\ 
\midrule
PR-UNK** (bin)   & 14k & 31.84$_{\pm 0.01}$ & 24.85$_{\pm 0.00}$ & 0.26\\ 
PR-BPE (bin)  &  69k & 31.77$_{\pm 0.01}$ & 24.55$_{\pm 0.01}$ & 0.20\\ 
PR-UNK** (cont)   & 9k & 31.85$_{\pm 0.02}$ & 24.85$_{\pm 0.01}$ & 0.26\\ 
PR-BPE (cont)  & 55k & 31.79$_{\pm 0.02}$ & 24.59$_{\pm 0.01}$ & 0.24\\ 
\midrule
SF-EL-UNK**  & 658k & 30.18$_{\pm 0.15}$ & 29.12$_{\pm 0.10}$ & 4.53\\ 
SF-EL-BPE & 590k  & 30.32$_{\pm 0.26}$  & 28.51$_{\pm 0.18}$ & 4.16\\ 
\midrule
BL-EL-UNK**  & 644k & 29.91$_{\pm 0.03}$ & 30.44$_{\pm 0.13}$ & 5.85\\ 
BL-EL-BPE  & 742k & 29.84$_{\pm 0.61}$ & 30.24$_{\pm 0.46}$ & \textbf{5.89} \\ 
\bottomrule
\end{tabular}
\subcaption{Domain adaptation from EP to TED.}

\end{subtable}%
}

\end{center}
\caption{Bandit NMT results (BLEU) on EP, NC and TED test sets. UNK* models involve UNK replacement only during testing, UNK** include UNK replacement already during training. For PR, either binary (bin) or continuous feedback (cont) was used. Control variates: average reward baseline (BL) and score function (SF). Results are averaged over two independent runs and standard deviation is given in subscripts. Improvements over respective out-of-domain models are given in the Diff.-columns.}
\label{tab:el}
\end{table*}

\subsection{Results}

In the following, we discuss the results of the experimental evaluation of the models described above. The out-of-domain baseline results are given in Table \ref{tab:nmtbaselines:out}, those for the in-domain baselines in \ref{tab:nmtbaselines:in}. The results for bandit learning on NC and TED are reported in Table \ref{tab:el}. For bandit learning we give mean improvements over the respective out-of-domain baselines in the Diff.-columns.

\paragraph{Baselines.}
The NMT out-of-domain baselines, reported in Table \ref{tab:nmtbaselines:out}, perform comparable to the linear baseline from \citet{SokolovETAL:16,SokolovETALnips:16} on NC, but the in-domain EP$\rightarrow$NC (Table \ref{tab:nmtbaselines:in}) baselines outperform the linear baseline by more than 3 BLEU points. Continuing training of a pre-trained out-of-domain model on a small amount of in domain data is very hence effective, whilst the performance of the models solely trained on small in-domain data is highly dependent on the size of this training data set. For TED, the in-domain dataset is almost four times as big as the NC training set, so the in-domain baselines perform better. This effect was previously observed by \citet{luong2015stanford} and \citet{FreitagETAL:16}. 

\paragraph{Bandit Learning.} 
The NMT bandit models that optimize the EL objective yield generally a much higher improvement over the out-of-domain models than the corresponding linear models: As listed in Table \ref{tab:el}, we find improvements of between 2.33 and 2.89 BLEU points on the NC domain, and between 4.18 and 5.18 BLEU points on the TED domain. In contrast, the linear models with sparse features and hypergraph re-decoding achieved a maximum improvement of 0.82 BLEU points on NC.

Optimization of the PR objective shows improvements of up to 1.79 BLEU points on NC (compared to 0.6 BLEU points for linear models), but no significant improvement on TED. The biggest impact of this variance reduction technique is a considerable speedup of training speed of 1 to 2 orders of magnitude compared to EL.

A beneficial side-effect of NMT learning from weak feedback is that the knowledge from the out-domain training is not simply ``overwritten''. This happens to full-information in-domain tuning where more than 4 BLEU points are lost in an evaluation on the out-domain data. On the contrary, the bandit learning models still achieve high results on the original domain. This is useful for conservative domain adaptation, where the performance of the models in the old domain is still relevant.

\paragraph{Unknown words.} By handling unknown words with UNK-Replace or BPEs, we find consistent improvements over the plain word-based models for all baselines and bandit learning models. 
We observe that the models with UNK replacement essentially benefit from passing through source tokens, and only marginally from lexical translations. Bandit learning models take particular advantage of UNK replacement when it is included already during training. The sub-word models achieve the overall highest improvement over the baselines, although sometimes generating nonsense words. 

\paragraph{Control variates.} Applying the score function control variate to EL optimization does not largely change learning speed or BLEU results. However, the average reward control variate leads to improvements of around 1 BLEU over the EL optimization without variance reduction on both domains.

\section{Conclusion}
In this paper, we showed how to lift structured prediction under bandit feedback from linear models to non-linear sequence-to-sequence learning using recurrent neural networks with attention. We introduced algorithms to train these models under numerical feedback to single output structures or under preference rankings over pairs of structures. In our experimental evaluation on the task of neural machine translation domain adaptation, we found relative improvements of up to 5.89 BLEU points over out-of-domain seed models, outperforming also linear bandit models. Furthermore, we argued that pairwise ranking under bandit feedback can be interpreted as a use of antithetic variates, and we showed how to include average reward and score function baselines as control variates for improved training speed and generalization. In future work, we would like to apply the presented non-linear bandit learners to other structured prediction tasks.

\section*{Acknowledgments}

This research was supported in part by the German research foundation (DFG), and in part by a research cooperation grant with the Amazon Development Center Germany.

\bibliography{bib}
\bibliographystyle{acl_natbib}

\end{document}